# Towards whole-body CT Bone Segmentation

André Klein[1,2], Jan Warszawski[2], Jens Hillengaß[3], Klaus H. Maier-Hein[1]

[1]Division of Medical Image Computing, Deutsches Krebsforschungszentrum (DKFZ)
[2]Medical Faculty, University of Heidelberg
[3]Section Multiple Myeloma, Department of Hematology, Oncology and
Rheumatology, University of Heidelberg
andre.klein@dkfz-heidelberg.de

**Abstract.** Bone segmentation from CT images is a task that has been worked on for decades. It is an important ingredient to several diagnostics or treatment planning approaches and relevant to various diseases. As high-quality manual and semi-automatic bone segmentation is very time-consuming, a reliable and fully automatic approach would be of great interest in many scenarios. In this publication, we propose a U-Net inspired architecture to address the task using Deep Learning. We evaluated the approach on whole-body CT scans of patients suffering from multiple myeloma. As the disease decomposes the bone, an accurate segmentation is of utmost importance for the evaluation of bone density, disease staging and localization of focal lesions. The method was evaluated on an in-house data-set of 6000 2D image slices taken from 15 whole-body CT scans, achieving a dice score of 0.96 and an IOU of 0.94.

## 1 Introduction

Fast and accurate automatic bone segmentation is important for analysis, staging and treatment planning of various diseases like multiple myeloma. Despite several years of research, it is still a significant challenge in some aspects [1] caused by the inhomogeneous structure and various shapes of bones and the fact that CT scans in clinical routine are often captured with a low dose which leads to inferior image quality.

Bones can be assigned to four different categories based on their shape: long bones, short bones, flat bones, and irregular bones [2]. As shown in Fig. 1 bones are composes of three different tissue types: cortical (compact) bone, cancellous (trabecular, spongy) bone and bone marrow. The cortical bone is the most dense and solid part with high Hounsfield Units (HU), surrounding the bone marrow compartment [2]. Because of the variation in density, the different types of bone have huge differences in HU. Cancellous bone and bone marrow are less dense, with HU being more similar to those of soft tissue like muscles. Pathological changes in the bone, e.g., caused by multiple myeloma can influence the density and therefore the HU of bone tissue [3].

The gold standard for bone segmentation is still semi-automated slice-by-slice hand contouring, which is very time-consuming [4]. Fully automatic bone segmentation has therefore been of great interest to research for a long time. Numerous studies can be found in literature addressing the issue, as described by Pinheiro et al. [5] and Buie et al. [5]. In the large number of studies, a lot of different approaches were proposed. Yet, bone segmentation is still considered an open problem in several aspects [1].

**Fig. 1.** CT scan of the femur. Cortical bone appears white and surrounds the less dense cancellous bone and the bone marrow.

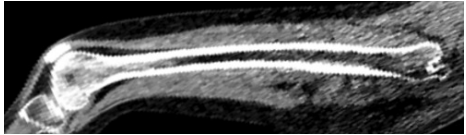

Some authors consider bone segmentation as a local problem, concentrating on specific bones. Krčah et al. and Younes et al. address the issue by focusing on the femur [1,6]. Younes et al. propose primitive shape recognition and statistical shape models. A more general approach is proposed by Pinheiro et al. [5]. They regard it as a local problem, too. However, they are not focusing on a particular bone, but on a user-defined region of interest, applying a level-set based protocol. Other authors like Pérez-Carrasco et al. apply more general heuristics to whole-body scans [7], in their case continuous max-flow optimization. Furthermore, approaches are based on region growing, intensity thresholding (e.g., Buie et al. [4]), energy minimizing spline curves, edge detection or combinations of these algorithms. They often rely on expensive pre- and post-processing steps or are depending on the specific initialization [8]. While deep learning algorithms have become a methodology of choice in many areas of automatic medical image segmentation problems [9], their performance on a bone segmentation task remains to be evaluated. Some initial work can be found in the "Bone Segmenter" project by Kevin Mader of 4Quant[1].

In this paper, we present our most recent efforts towards bone segmentation on whole-body CT images, more specifically: low quality low dose CT scans that were captured as part of a PET/CT study during standard assessment for patients with multiple myeloma. We propose a network based on the U-Net architecture by Ronneberger et al. [10]. The goal of our work is to locate and segment cortical and cancellous tissue as well as bone marrow of long, short, flat and irregular bones in whole-body scans of patients with multiple myeloma.

## 2   Materials and methods

### 2.1   Data

We use an in-house data-set that consists of 15 whole-body low quality CT scans of patients diagnosed with multiple myeloma. We perform a k-fold cross-validation with k=5, thus using 9 patients for training ($\approx$3800 slices), 3 for validation ($\approx$1300 slices) and 3 for testing ($\approx$1300 slices) in each fold. Slices are 512x512 pixels, and each scan has between 380 and 450 slices. All datasets have an equal spacing of 0.98x0.98x4 mm³. The ground truth segmentation has been performed by a medical expert who was provided with segmentations generated by an intensity threshold. Slices were corrected using the segmentation plugin of the Medical Imaging Interaction Toolkit (MITK) [11].

---

[1] https://github.com/4Quant/Bone-Segmenter

## 2.2 Architecture

We adapt the U-Net architecture that was initially proposed by Ronneberger et al. [10] as shown in Fig. 2.2. The U-Net is a fully convolutional network with 18 convolutional layers. It consists of a downsampling and a symmetric upsampling path and uses skip connections to fast forward features from shallow layers to deep ones. Our model uses padded convolutions with a kernel size of 3 to keep the spatial output dimensions equal to the input. We resized the layers to match our image size of 512x512 pixels. The number of feature channels in the first convolutional channel is set to 64 as proposed in the original paper and doubled whenever the network increases in depth. We use a 2D architecture and provide the network with axial slices as input images.

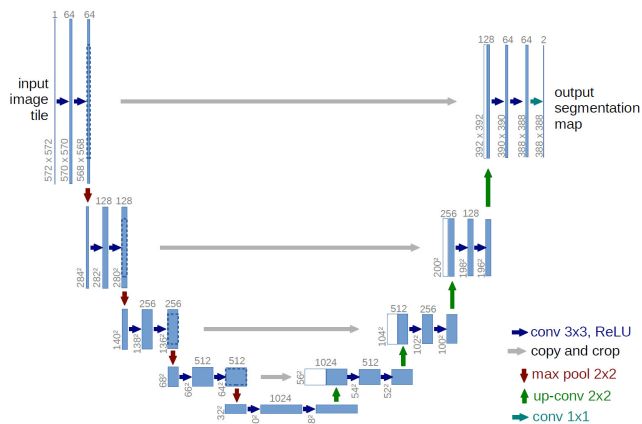

**Fig. 2.** U-Net architecture as proposed by Ronneberger et al. [10]. The architecture consists of a contracting path that captures semantic information and a symmetric expanding path that enables precise localization information [10].

## 2.3 Training

Data augmentation is used to more efficiently train our network given the amount of training data [10]. We make use of ±180° rotations around the axial axis, as well as randomly mirroring in x- and y-direction. We use a categorical cross-entropy loss and an adam optimizer with a learning rate of 0.0005, $\beta_1 = 0.5$, $\beta_2 = 0.999$ for our training. Our network is trained for 60 epochs with batch size 8 and 500 batches per epoch.

## 3    Results

The proposed segmentation algorithm achieved a dice score of 0.96±0.02 and an intersection over union (IOU) of 0.94±0.02. In comparison, the standard procedure, i.e. thresholding + morphological operations, achieved a dice score of 0.85±0.04 and an IOU of 0.78±0.06.

Both, the proposed and the standard procedure, worked well for cortical bone due to its high HU values. An example is shown in Fig. 3 for the proposed and in Fig. 4 for the standard procedure. The main issues arose when segmenting bone marrow and spongy bone. As expected, the standard approach did not segmented these structures well and also often mistook the table in the images as bony structure (see Fig. 4). These issues are partly solved by our approach. However, performance on more complex body regions like the chest was still challenging as the network tends to oversegment bone like tissue such as cartilage. The most difficult task for both approaches were patients with hip or knee replacement. Segmenting bone on the according slices is a difficult task because of the artifacts that have similar HUs as cortical bone, and they are not represented sufficiently frequent in our data set for the proposed method to learn how to adequately handle such situations (see Fig. 5).

**Fig. 3.** The network performs best on long bones like the femur.

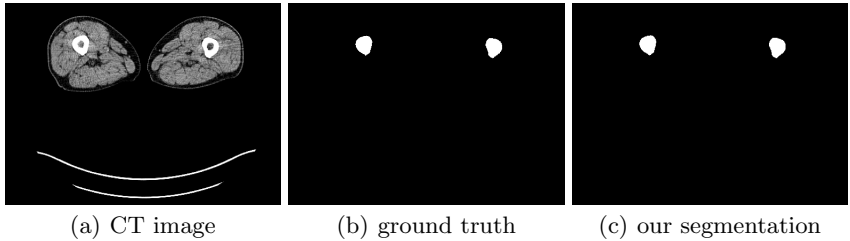

(a) CT image             (b) ground truth             (c) our segmentation

The segmentation of a whole-body CT scan (512x512x400) took about 30s on an NVIDIA Titan X GPU.

## 4    Discussion

In this paper, we present a deep learning approach for the simultaneous segmentation of long, short, flat and irregular bones including cortical, cancellous and bone marrow structures. Our network achieved promising dice and IOU scores despite the low image quality of the applied whole-body CT scans. As expected, the segmentation of smaller bones like the ribs was more challenging, which is probably related to the fact that only small pieces of each bone are visible on each slice and that tey are surrounded by more complex tissue combinations (see Fig. 6).

**Fig. 4.** Segmentation created with threshold + morphological operations. The bone marrow is not segmented and the table is mistaken with bone.

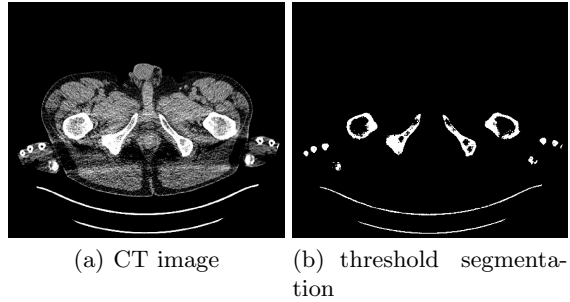

(a) CT image

(b) threshold segmentation

**Fig. 5.** Artefacts caused by tooth crowns or artificial joints lead to imperfect segmentations.

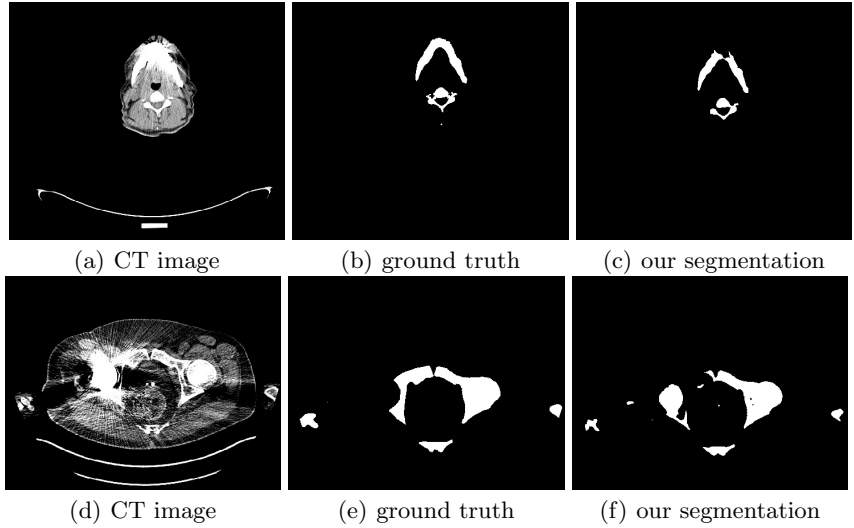

(a) CT image          (b) ground truth          (c) our segmentation

(d) CT image          (e) ground truth          (f) our segmentation

Many different bone segmentation approaches have been published so far. It is not easy to provide a fair comparison of the different algorithms, as a lot of the work is focused on restricted problems like the segmentation of specific bony structures. To our knowledge, there do currently not exist any public benchmark datasets for the problem of general bone segmentation. Pérez-Carrasco et al. [7] presented a solution that used a continuous max-flow optimization to segment CT images. We did not reimplement the method, but on their in-house dataset the authors achieved a dice score of 0.91, requiring approx. 0.5s processing time per slice (512x512 pixels).

Our network was trained on images from a single scanner only. A larger dataset with higher heterogeneity could be established in the future to establish a more general bone segmentation method that applies to a variety of scanners

**Fig. 6.** Segmentation of smaller bones like rips is a harder task but still provides good results.

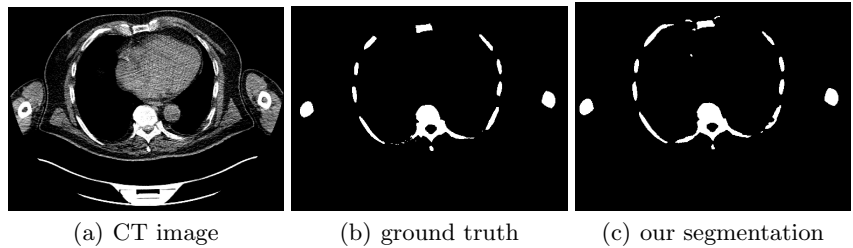

(a) CT image          (b) ground truth          (c) our segmentation

and different levels of image quality. We will continue to further expand our reference dataset and plan to develop semi-supervised approaches that leverage unlabeled input data during learning.